%% file: 00_main.tex
\documentclass[11pt,a4paper]{article}
\usepackage[hyperref]{acl2021}
\usepackage{times}
\usepackage{latexsym}
\usepackage{graphicx}
\usepackage{amsmath}
\usepackage{paralist}
\graphicspath{ {./figures/} }

\usepackage{tabularx}
\usepackage{boldline}
\usepackage{booktabs}
\usepackage{multirow}
\usepackage{xspace}
\usepackage{makecell}

\usepackage{arydshln}
\usepackage{tabu}
\newcommand{\para}[1]{\vskip 1mm\noindent\textbf{#1}~~}

\usepackage{microtype}

\aclfinalcopy 

\usepackage{kotex}

\title{Back-Translated Task-Adaptive Pretraining: \\
Improving Accuracy and Robustness on Text Classification}

\author{
    Junghoon Lee \quad
	Jounghee Kim \quad
	Pilsung Kang\thanks{\ \  Corresponding author.} \\\\ 
	Korea University, Seoul, Republic of Korea \\ 
	{\tt \{junghoon\_lee, jounghee\_kim, pilsung\_kang\}@korea.ac.kr}
}
\date{}

\begin{document}
\maketitle
\input{0_abstract.tex}
\input{1_introduction.tex}
\input{2_background.tex}
\input{3_bt-tapt.tex}
\input{4_experiments}

\input{5_robustness_noise}
\input{7_conclusion}



\bibliographystyle{acl_natbib}
\bibliography{reference}

\end{document}

%% file: 0_abstract.tex
\begin{abstract}
Language models (LMs) pretrained on a large text corpus and fine-tuned on a downstream task becomes a \textit{de facto} training strategy for several natural language processing (NLP) tasks. Recently, an adaptive pretraining method retraining the pretrained language model with task-relevant data has shown significant performance improvements. However, current adaptive pretraining methods suffer from underfitting on the task distribution owing to a relatively small amount of data to re-pretrain the LM. To completely use the concept of adaptive pretraining, we propose a \textit{back-translated task-adaptive pretraining} (\textbf{BT-TAPT}) method that increases the amount of task-specific data for LM re-pretraining by augmenting the task data using back-translation to generalize the  LM to the target task domain. The experimental results show that the proposed \textbf{BT-TAPT} yields improved classification accuracy on both low- and high-resource data and better robustness to noise than the conventional adaptive pretraining method.
\end{abstract}

%% file: 1_introduction.tex
\section{Introduction}

In the history of natural language processing (NLP), the rise of large language models (LMs) trained on a huge amount of text corpora was a game-changer. Before the advent of these LMs, an NLP task-specific model was trained only on a small amount of labeled data. Due to the high cost of label annotation for text data, insufficient training data was always one of the main obstacles to NLP model progress \cite{liu2016recurrent}. However, researchers found that large LMs trained on a huge amount of unlabeled, i.e., task independent, text data, such as {\textsc{BERT}\xspace} \cite{devlin2019bert} or {\textsc{GPT-3}\xspace} \cite{brown2020language}, significantly improved the performance of various NLP tasks just by beginning with a pre-trained LM and fine-tuning it using the task-specific labeled data. This strategy, i.e., combining a large pretrained LM with task-specific fine-tuning, outperformed the state-of-the-art models in various NLP tasks such as text classification \cite{howard2018universal}, natural language inference \cite{peters2018deep}, summarization \cite{lewis-etal-2020-bart}, and question answering \cite{howard2018universal, lan2019albert}. 

Although pretrained LMs have generalized language representations based on the corpora collected from a wide range of domains, it is not sufficient to learn completely a specific domain of some downstream tasks. To overcome this limitation, adaptive pretraining that re-pretrains the pretrained LM with the task-relevant data before fine-tuning phase was proposed \cite{beltagy2019scibert, sun2019fine, lee2020biobert, gururangan2020don}. In practice, however, it is challenging to obtain another in-domain data that share the same characteristics with the task data due to the scarcity of the tasks, such as developing an intent classifier using data collected from a newly-launched chatbot \cite{anaby2020not}. Moreover, when only a few task data is available, adaptive pretraining on the task data is insufficient to generalize the LM to the task distribution. Hence, the task-adaptively pretrained LMs might still be underfitted on the task distribution even though one can achieve better performance by employing adaptive pretraining.

To solve this problem, we propose a \textit{back-translated task-adaptive pretraining} (\textbf{BT-TAPT}) strategy that augments the task data based on back-translation to secure more amount of task-relevant data to better generalize the pretrained LMs to the target task domain. Although text augmentation has helped improve the generalization and robustness of NLP models in various tasks such as classification, translation, and question answering \cite{sennrich2016neural, edunov2018understanding, yu2018qanet, wei2019eda, xie2019unsupervised}, the augmented data are only used in the fine-tuning step thus far.

\begin{figure}[t]
\includegraphics[width=\columnwidth]{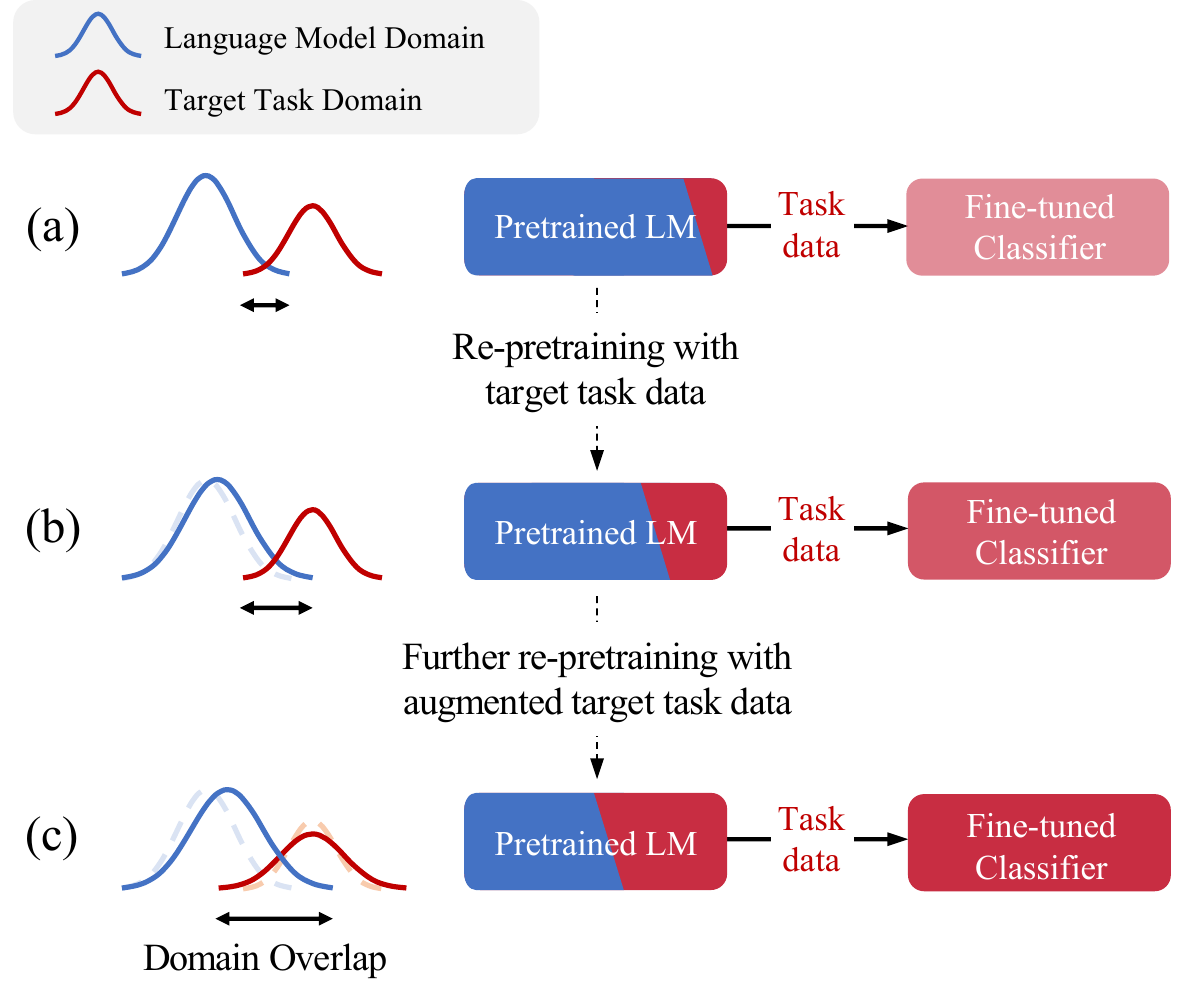}
\caption
{
Comparison between the existing language model pretraining and the proposed methods.
(a) General pretraining.
(b) Adaptive pretraining with task-relevant data.
(c) Proposed method -- further re-pretraining with back-translated task data.
}
\label{fig:1}
\centering
\end{figure}

Figure \ref{fig:1} illustrates the expected advantage of the proposed \textbf{BT-TAPT}. In \textbf{BT-TAPT}, an adaptive pretraining of LM is first conducted with the original task data. The task corpus is then augmented based on the back-translation technique using an appropriate sampling method such as nucleus sampling \cite{holtzman2019curious}. In this way, we can generate various paraphrases from the original task corpus. These augmented task corpora are used to re-pretrain the adaptively pretrained LM again to better generalize the LM for the target task domain. Based on these consecutive pretraining procedures, we can expect that the overlap between the language model domain and the target task domain would increase as described in Figure \ref{fig:1}.

To verify the proposed \textbf{BT-TAPT}, we employed two well-known pretrained LMs: {\textsc{BERT}\xspace} and {\textsc{RoBERTa}\xspace} \cite{liu2019roberta}. The performance of \textbf{BT-TAPT} was evaluated on six text classification datasets and compared with two benchmark methods: pretrained LM and task-adaptive pretrained LM. In general, the experimental results show that the proposed \textbf{BT-TAPT} yields higher classification accuracy than the benchmark methods. In addition, we verified the robustness of \textbf{BT-TAPT} by generating five types of noise for the test dataset and comparing the performance with the benchmark methods. As expected, \textbf{BT-TAPT} showed more robust classification performance than the baseline methods, supporting that the back-translation-based augmented data improves the generalization ability of the pretrained LMs.

Consequently, our contribution can be summarized as follows:
\begin{compactitem}
    \item We propose a new adaptive pretraining method (\textbf{BT-TAPT}) that can generalize LMs to task distribution using back-translation-based augmented downstream task corpus.
    \item \textbf{BT-TAPT} enhances the performance of downstream tasks.
    \item \textbf{BT-TAPT} shows better robustness to noisy text data.
\end{compactitem}

%% file: 2_background.tex
\section{Background}
This section briefly reviews three key components of the study: masked language model, language model adaptation, and text data augmentation.

\subsection{Masked Language Model}
Masked language model (MLM) proposed in \citet{devlin2019bert} is a pretraining method that predicts original tokens in a sentence where some tokens are masked with a special token \texttt{[MASK]}. 
Let $X=(x^{(1)}, x^{(2)}, ..., x^{(N)})$ denote a set of unannotated sentences, where $x=(t_1, t_2, ... t_M)$ is a sequence of tokens in a sentence, and $t_i \in x$ is a token in a sequence. In the pretraining process of the masked language model, noise is added to the sequences $x$ by randomly replacing some tokens with \texttt{[MASK]} token. Let $\hat{x} \in \hat{X}$ be a noised sentence and $\bar{x} \in \bar{X}$ be a subset of tokens that is randomly replaced. The training objective can be formulated as follows:
\begin{equation*}
\displaystyle \min_{\theta} L_{MLM}(\theta;\hat{X},\bar{X})=-\sum_{\hat{x}, \bar{x} \in \hat{X}, \bar{X}}log p_{\theta}(\bar{x}|\hat{x}).
\end{equation*}
Because MLMs do not require any task-specific labels, they can be trained on a large unannotated corpus, which makes researchers believe they can learn a general representation of single or even multiple languages. 
This belief has been partially supported because pretrained MLMs based on a gigantic text corpus followed by fine-tuning on a small task-specific data often outperform the state-of-the-art models for a wide range of NLP tasks \cite{liu2019roberta, lan2019albert, song2019mass, raffel2020exploring}. 



\subsection{Language Model Adaptation}
LMs are pretrained with a vast amount of corpora from different domains such as \textsc{BookCorpus} \cite{zhu2015aligning}, \textsc{Wikipedia}, \textsc{CC-News} \cite{liu2019roberta}, \textsc{OpenWebText} \cite{Gokaslan2019OpenWeb}, and \textsc{Stories} \cite{trinh2018simple}. Contrary to the expectation that LMs can properly generalize the language representation despite the text corpora based on which the LMs are trained, LMs have been reported that they are highly dependent on the training data domain. Moreover, their performances on other domains are not as good as on the training data domain. 

To resolve this issue, language model adaptation, which re-pretrain the LMs before fine-tuning for specific tasks \cite{sun2019fine, gururangan2020don}, was proposed. There are two main streams in language model adaptation: \textit{domain-adaptive pretraining} (\textbf{DAPT}) and \textit{task-adaptive pretraining} (\textbf{TAPT}). \textbf{DAPT} re-pretrains the LM based on a new dataset on the same domain of a target task, whereas \textbf{TAPT} directly re-pretrain the LM using the given target task. If a task domain is uncommon or the data acquisition cost is expensive, \textbf{TAPT} can be more suitable than \textbf{DAPT}.

\subsection{Text Data Augmentation}
Data augmentation for NLP is less actively used than image augmentation thus far. While many simple label-unchanged operations exist for images such as flipping, rotation, cropping, and translation, languages barely have such operations. Hence, synonym replacement using WordNet has been the most popular augmentation method in NLP \cite{wang2015s, zhang2015character}. 

Recently, easy data augmentation (EDA) techniques consisting of synonym replacement, random insertion, random swap, and random deletion proved useful for text classification \cite{wei2019eda}. Despite the performance enhancement, the EDA has a significant limitation that the semantics of the original sentences are usually damaged except the synonym replacement \cite{kumar2020data}. 
As an alternative to preserving the semantics of original sentences, back-translation, translating a sentence $\mathbf{x}_{source}$ in the source language into the other language and then translates it back into $\hat{\mathbf{x}}_{source}$ in the source language, was proposed. The technique showed to improve the performances in various NLP tasks, such as machine translation \cite{sennrich2016improving}, question answering \cite{yu2018qanet}, and semi-supervised text classification \cite{xie2019unsupervised}. 

\begin{figure}[t!]
\includegraphics[width=\columnwidth]{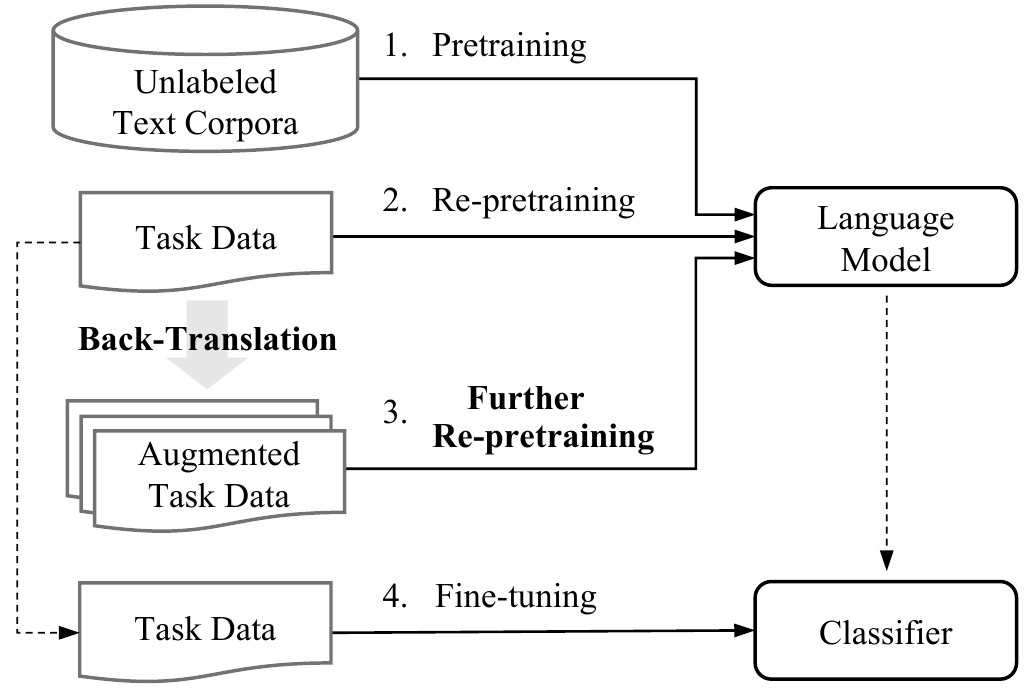}
\caption
{
Process of \textbf{BT-TAPT}. With the pretrained LM, first, we re-pretrained it with the task data. Subsequently, we further re-pretrained it using augmented task data based on back-translation. Finally, the further re-pretrained LM is fine-tuned using the original task data.
}
\label{fig:2}
\centering
\end{figure}



%% file: 3_bt-tapt.tex
\input{tables/1.data_description}

\section{Proposed method}
We introduce a \textit{back-translated task-adaptive pretraining} (\textbf{BT-TAPT}), a new adaptive pretraining strategy that helps adaptive pretraining when task data is insufficient, and in-domain data is unavailable. The overall process of \textbf{BT-TAPT} is shown in Figure \ref{fig:2}.

\subsection{Task Data is Insufficient}
While using \textbf{TAPT} contributes to task-specific performance improvement, \citet{gururangan2020don} showed that continued pretraining using human-curated unlabeled data -- corpus from the same source with task data -- ensured additional performance gain, which is named human-curated \textbf{TAPT} Simultaneously, the automatic selection approach retrieving unlabeled data aligning with the task data distribution from the in-domain corpus is also proven beneficial. Despite the favorable results, using those methods is impossible when human-curated data or in-domain data are unavailable. Motivated by this limitation, we propose an advanced adaptive pretraining approach requiring only task data.


\subsection{Back-Translated TAPT}
If the amount of task data is insufficient during \textbf{TAPT}, the LM may still be underfitted on the task domain. In this case, the LM would be more generalized to the task if more task-related sentences were available. The proposed \textbf{BT-TAPT} is an additional adaptation method using human-like task-related sentences. To make plausible sentences, we use a back-translation using a nucleus(top-$p$) sampling. We used the nucleus sampling instead of traditional beam search because the former can better implement the purpose of augmentation than the latter, creating label-unchanged and semantic-preserved data that are not exactly the same as the original data. Although back-translated sentence using beam search tends to be an almost identical sentence to the original sentence, sentences generated by back-translation using the proper sampling method are usually paraphrases of the original sentence, which have various expressions   without significantly deviating from the domain of the original data. 

When the LM is further pretrained only with task data, noises applied to the input at every epoch are changing the location of the \texttt{[MASK]} tokens and infinitesimal random word transition.
Otherwise, noise used in \textbf{BT-TAPT} includes not only the change of \texttt{[MASK]} positions or word transition but also synonym replacement or rephrasing facilitated by the sampling-based back-translation.

\subsection{BT-TAPT Process}
We first apply \textbf{TAPT} on the pretrained LM to expose it to the task domain sentence with correct grammar. Then, we generate multiple sentences with back-translation. We adopt nucleus sampling with $p=0.95$ to ensure proper diversity and fluency for the generated sentences. We generate 20 sentences for each original sentence to expose the LM to semantically and syntactically sufficiently diverse expressions. We further re-pretrain the LM with those sentences where domain-related but less accurate expressions appear. Learning from those paraphrases, the LM can cover a broader range of task distribution, as depicted in figure \ref{fig:1}. After the entire adaptive pretraining phase is completed, the LM is fine-tuned on the task data.

%% file: tables/1.data_description.tex
\begin{table*}[t]
    \centering
    \small
    \begin{tabular}{lcccccccc}
    \toprule
         \multirow{2}*{Data} & \multirow{2}*{Task} & \multirow{2}*{\# Class} & Average & Maximum & \multirow{2}*{\# Train} & \multirow{2}*{\# Train (Low.)} & \multirow{2}*{\# Dev} & \multirow{2}*{\# Test} \\
         ~ & ~ & ~ & lengths & lengths & ~ & ~ & ~ & ~ \\
         \midrule
         IMDB & \small Sentiment & 2 & 297 & 2,291 & 20,000 & 5,000 & 5,000 & 25,000 \\
         MR & \small Sentiment & 2 & 28 & 73 & 5,000 & - & 2,662 & 3,000 \\
         SST2 & \small Sentiment & 2 & 25 & 70 & 6,920 & - & 872 & 1,822 \\
         \textsc{Amazon} & \small Sentiment & 2 & 202 & 7,810 & 115,251 & 11,525 & 5,000 & 25,000 \\
         TREC & \small Question & 6 & 13 & 38 & 4,906 & - & 546 & 500 \\
         \textsc{AGNews} & \small Topic & 4 & 52 & 255 & 115,000 & 10,000 & 5,000 & 7,600 \\ 
    \bottomrule
    \end{tabular}{\parfillskip=0pt\par}
    \caption{Description of six classification benchmark datasets. \# Train (Low.) refers to a number of data downsampled from the original training data for low-resource scenario, following the low-resource setting in \citet{gururangan2020don}}
    \label{tab:1}
\end{table*}

%% file: 4_experiments.tex
\input{tables/2.exp_low}

\input{tables/3.exp_high}

\section{Experiments}
\label{sec:exp}

We verify the proposed \textbf{BT-TAPT} based on six widely-studied classification datasets with two well-known pretrained LMs: BERT and \textsc{RoBERTa}\xspace. We compare the performances of \textbf{BT-TAPT} with two benchmark methods: base pretrained model and \textbf{TAPT}.

\subsection{Datasets \& Performance Metrics}
Four sentiment classification datasets, i.e., IMDB \cite{maas2011learning}, MR \cite{pang2005seeing}, SST2 \cite{socher2013parsing}, and \textsc{Amazon} \cite{he2016ups, gururangan2020don} and two other classification datasets, i.e., 
TREC \cite{li2002learning} question classification and \textsc{AGNsews} \cite{zhang2015character} topic classification, were used in the experiments. As a classification performance metric, a simple accuracy was used except for \textsc{Amazon}, for which the macro-$F_1$ is used owing to the class imbalance.


We divided the datasets into two groups: datasets with more than 20,000 training examples were considered high-resource datasets, while the others were considered low-resource datasets. To formulate additional low-resource datasets, IMDB and \textsc{AGNews} were down-sampled 2,500 per class, and \textsc{Amazon} was down-sampled 10\% of the training dataset. The description of each dataset is summarized in Table \ref{tab:1}.

\subsection{Training Details}
As the pretrained LMs, BERT-base and \textsc{RoBERTa}\xspace-base trained from huggingface \cite{wolf-etal-2020-transformers} \footnote[1]{ https://github.com/huggingface/transformers} were employed. For back-translation, we employed the transformer-big model of Facebook \cite{ng2019facebook} \footnote[2]{ https://github.com/pytorch/fairseq/tree/master/examples/translation}, a WMT'19 winner. We translated sentences from English to German and then back-translated them from German to English. When generating the translated sentence in the decoding process, we used the nucleus sampling with the sampling probability of $p=0.95$. The back-translation generated a total of 20 sentences for each original sentence. 

For both \textbf{TAPT} and \textbf{BT-TAPT}, we re-pretrained the pretrained LMs 100K steps in the high-resource setting (IMDB, \textsc{Amazon}, and \textsc{AGNews}), following \citet{sun2019fine}.
In contrast, in the low-resource setting, we re-pretrained 50K steps because the model converged faster with a small dataset than a large dataset. 
We used a batch size of 64, with a max sequence length of 512, and AdamW optimizer using a learning rate of 5e-5 with a linear learning rate scheduler for which the 10\% of the steps were used for warming-up, and the rest of steps were used for decay. 
For task-specific fine-tuning, the modes were trained 2 or 3 epochs with a batch size of 8, 16, or 32, a weight decay of 0 or 0.01, and a fixed learning rate of 2e-5 with the same linear learning rate scheduler as re-pretraining. 
We stopped the fine-tuning when the validation loss starts to increase and reported the best performance among all hyper-parameter combinations.
To compensate for the effect of random seed initialization, we repeated each experiment five times and reported the average and the standard deviation of the performance metrics. All the experiments were conducted using the Nvidia DGX-station with 4 $\times$ 32GB Nvidia V100 GPUs.

\begin{figure}[t!]
\includegraphics[width=\columnwidth]{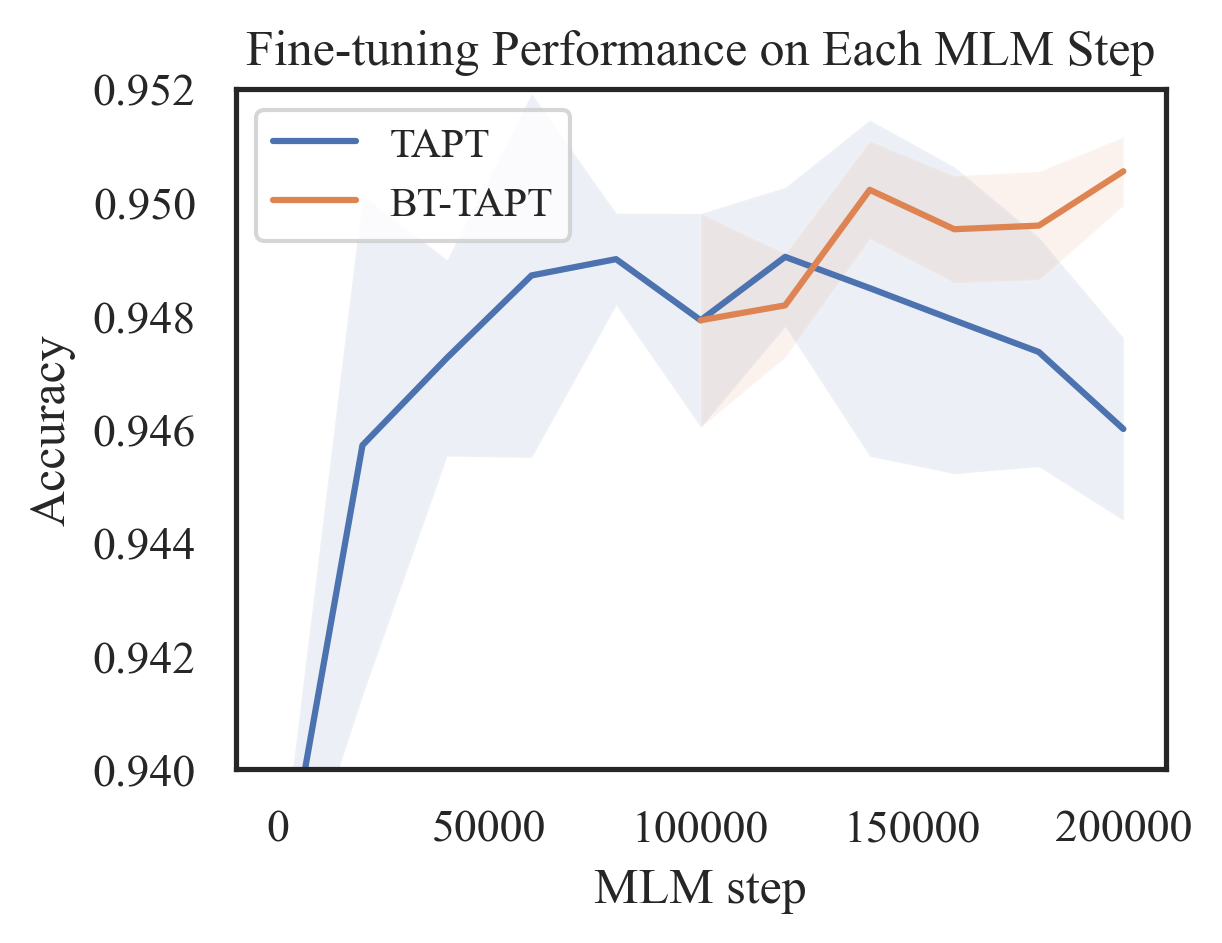}
\caption
{
Classification accuracy (solid line) with the standard deviation of five repetitions (shaded area) for the IMDB test dataset with the BERT LM.}
\label{fig:4}
\centering
\end{figure}

\subsection{Text Classification Performance}
Tables \ref{tab:2} and \ref{tab:3} show the classification performance of each dataset in the high- and low-resource settings, respectively. 
In both tables, we can observe that further re-pretraining with the augmented dataset based on the back-translation improves the model either in terms of classification accuracy or the model stability regardless of the base pretraind LM. The proposed \textbf{BT-TAPT} yields either higher classification accuracy without the increase in standard deviation than \textbf{TAPT} (e.g., IMDB with BERT, SST2 with BERT, and MR with \textsc{RoBERTa}) or reduce the performance variation while maintaining the classification accuracy (e.g., IMDB with \textsc{RoBERTa}, \textsc{AGNews} with \textsc{RoBERTa}, SST2 with \textsc{RoBERTa}). Note that there are some exceptions: re-pretraining might harm the performance as mentioned in \citet{sun2019fine} for those datasets with insufficient amount of sentences to generally represent the domain distribution, such as TREC dataset with \textsc{RoBERTa} and MR dataset with BERT.
However, \textbf{BT-TAPT} seems to succeed to compensate the gap by supplementing additional augmented data.

\input{tables/4.change_augmentation}
Figure \ref{fig:4} shows the effect of the re-pretraining of LMs with task data and back-translated task data in the full IMDB dataset with the BERT model. The solid line denotes the classification performance after the fine-tuning is completed using the re-pretrained LMs with the corresponding training steps in the $x$-axis, and the shaded area is the standard deviation of accuracy for the five repetitions. With \textbf{TAPT}, the accuracy is increased along with the step size. However, the effect of \textbf{TAPT} vanishes beyond a certain step size. For example, the accuracy degenerates after 80K in Figure \ref{fig:4}. Note also that the performance variation is also unstable with \textbf{TAPT}. When the proposed \textbf{BT-TAPT} is applied after 100K steps, the accuracy is rebounding along with the training steps. We can also observe that the accuracy not only further improves but also retains a low variability with \textbf{BT-TAPT} because a narrow bandwidth of the shaded area is retained.

\subsection{Comparing Augmentation Methods}
We compared back-translation with other widely used augmentation methods -- EDA, Embedding, and TF-IDF on adaptive pretraining. 
Embedding \cite{mrkvsic2016counter} replaces an arbitrary token with a close token in the embedding space. In contrast, TF-IDF \cite{xie2019unsupervised} replaces uninformative words with low TF-IDF scores in sentences while preserving those with high TF-IDF scores. As a baseline, we also include the model without any augmentation method, which means the \textbf{TAPT} is applied again with the same training steps for the other augmentation methods.

We first applied \textbf{TAPT} on BERT-base for 50K steps using the IMDB low-resource dataset and then re-pretrained using the augmented dataset generated from each method using a transformation probability of 0.1. Table \ref{tab:4} shows the classification accuracy for different augmentation methods. As mentioned regarding Figure \ref{fig:4}, excessive \textbf{TAPT} often degenerated the final performance: another 100K steps of \textbf{TAPT} resulted in a $0.2\%p$ lower accuracy than that without additional re-pretraining. In addition, none of the benchmark text augmentation methods can neither succeed in improving the classification performance nor reduce the performance variation. Only the proposed \textbf{BT-TAPT} enhanced the classification accuracy without the loss of variation. The reason behind this observation is that the back-translation can generate diverse and realistic paraphrases, while the others tend to modify the sentence partially with a simple technique. These simply modified sentences sometimes harm the semantic meaning of the original sentence, which disturbs an appropriate LM re-pretraining.



\begin{figure}[t!]
\centering{\includegraphics[width=0.85\columnwidth]{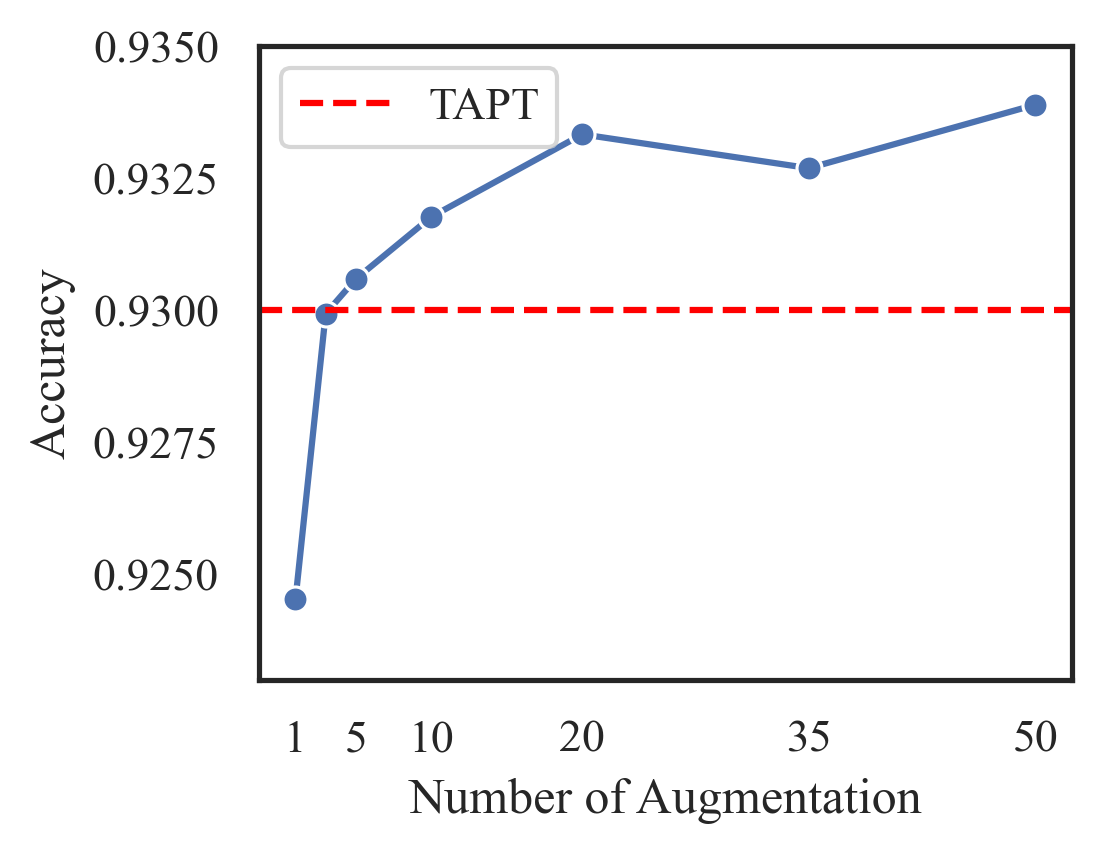}}
\caption
{
Accuracy of BERT fine-tuned on IMDB low-resource test dataset after applying \textbf{BT-TAPT} using different numbers of back-translated paraphrase per original sentence.}
\label{fig:6}
\centering
\end{figure}

\subsection{Quantity of Augmentation}
To investigate the number of back-translation for the downstream task performance, we generated between 1 to 50 sentences in the \textbf{BT-TAPT} process from the IMDB low-resource dataset. 
Figure \ref{fig:6} shows the classification accuracy regarding the number of augmentations. In general, classification accuracy increases with \textbf{TAPT} as the number of back-translated sentences increases and becomes mature beyond a certain number of augmentations. Because the performance differences beyond 20 augmentations are marginal, we chose 20 as the final number of back-translated augmentations per sentence.

\begin{figure}[t!]
\centering{\includegraphics[width=0.8\columnwidth]{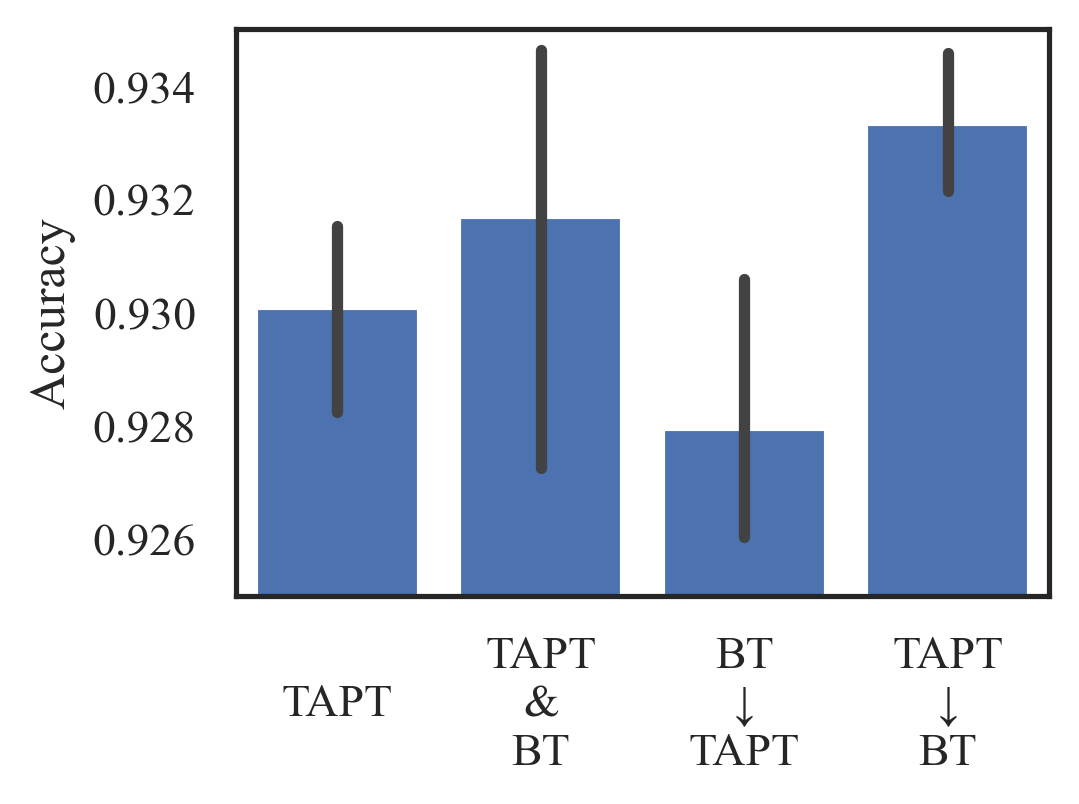}}
\caption
{
Accuracy of fine-tuned BERT with different order of \textbf{TAPT} and back-translation (\textbf{BT}) in \textbf{BT-TAPT} for IMDB low-resource test dataset.}
\label{fig:7}
\centering
\end{figure}

\subsection{Comparison of BT-TAPT Strategies}
We investigated all possible strategies of back-translation deployment in \textbf{BT-TAPT}. \textbf{TAPT \& BT} implies using both task and back-translated data for re-pretraining simultaneously. \textbf{BT → TAPT} and \textbf{TAPT → BT} refer to the re-pretraining with back-translated data and further re-pretraining with task data sequentially and vice versa. As seen in Figure \ref{fig:7}, applying \textbf{TAPT} followed by re-pretraining with back-translated sentences yields the highest accuracy with the smallest variation. However, simultaneously using the back-translated sentences on re-pretraining enlarges the deviation even though it improves the accuracy, which implies that the language model should adjust to the target domain with a well-formed corpus first and then be exposed to various domain-related augmented sentences.

\begin{figure}[t!]
\centering{\includegraphics[width=1.0\columnwidth]{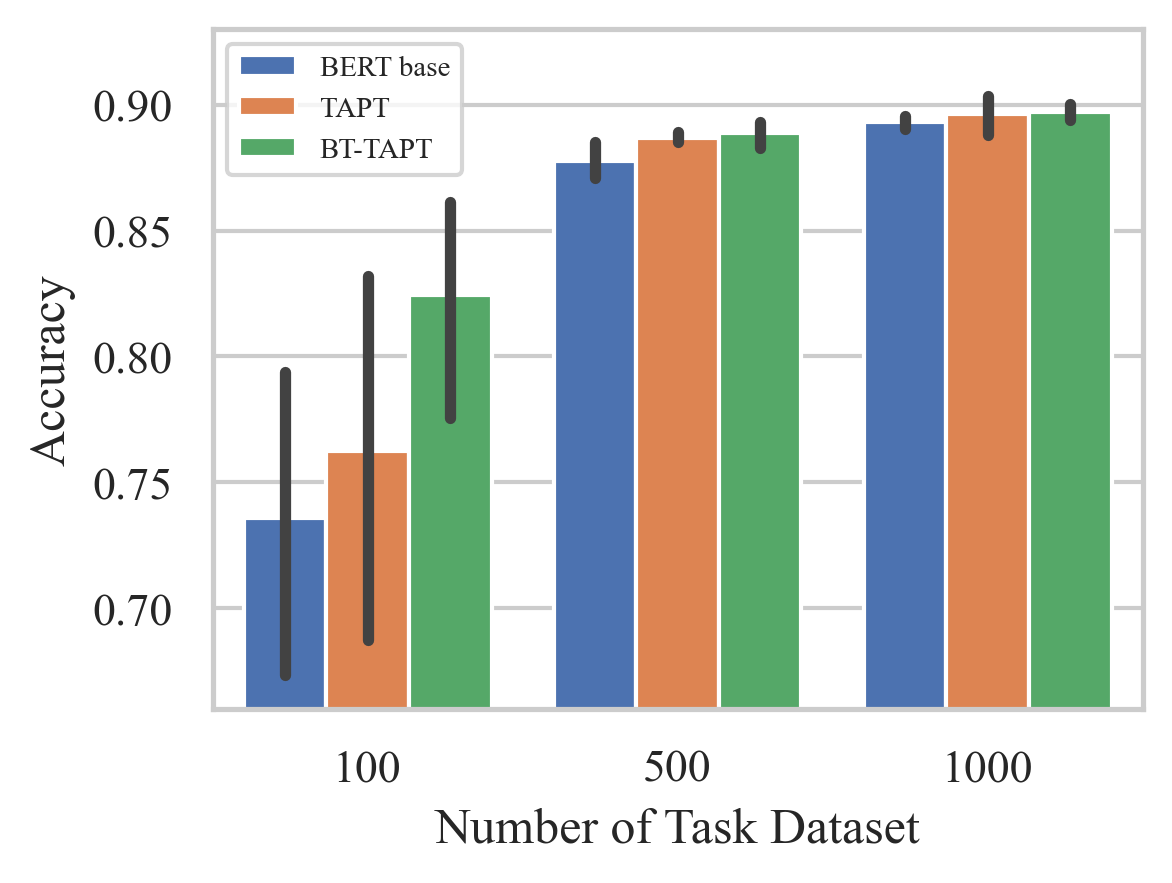}}
\caption
{
Test accuracy of BERT fine-tuned on the varying amount of task data. The indicated quantity of IMDB dataset were used in \textbf{TAPT}, \textbf{BT-TAPT}, and fine-tuning.}
\label{fig:8}
\centering
\end{figure}

\subsection{Performance on Small Datasets}
In a real-world situation, the task data is commonly insufficient. To validate that the \textbf{BT-TAPT} can effectively handle the data shortage, we conducted additional experiments with only 100, 500, and 1,000 sentences of the IMDB dataset. As shown in Figure \ref{fig:8}, \textbf{BT-TAPT} is found to be much beneficial under an extremely low-resource circumstance because the performance is noticeably improved with only 100 sentences. Thus, this result supports that expression-diversified but semantic-preserved augmented texts based on back-translation can help LMs adapt to the domain distribution.

%% file: tables/2.exp_low.tex
\begin{table*}[t]
    \centering
    \small
    \begin{tabular}{lcccccc}
    \toprule 
         & IMDB & \textsc{Amazon} & \textsc{AGNews} & TREC & MR & SST2 \\ 
         \midrule[0.05em]
         BERT & 92.2\textsubscript{0.3} & 60.8\textsubscript{2.3} & 92.1\textsubscript{0.1} & 96.7\textsubscript{0.1} & \textbf{86.5\textsubscript{0.7}} & 91.0\textsubscript{0.7} \\
         + \textsc{tapt} & 93.0\textsubscript{0.2} & 67.0\textsubscript{0.8} & \textbf{92.7\textsubscript{0.1}} & 96.0\textsubscript{0.3} & 85.7\textsubscript{0.3} & 90.6\textsubscript{0.3} \\
         + \textsc{bt-tapt} & \textbf{93.3\textsubscript{0.2}} & \textbf{67.3\textsubscript{0.9}} & \textbf{92.7\textsubscript{0.1}} & \textbf{96.9\textsubscript{0.3}} & 86.4\textsubscript{0.4} & \textbf{92.4\textsubscript{0.3}} \\ 
         \midrule[0.05em]
         \textsc{RoBERTa}\xspace & 94.2\textsubscript{0.3} & 63.7\textsubscript{1.9} & 92.4\textsubscript{0.3} & \textbf{96.7\textsubscript{0.5}} & 89.3\textsubscript{0.7} & 93.5\textsubscript{0.6} \\
         + \textsc{tapt} & \textbf{94.4\textsubscript{0.2}} & 64.5\textsubscript{2.8} & \textbf{92.6\textsubscript{0.3}} & 96.2\textsubscript{0.2} & 89.4\textsubscript{0.4} & \textbf{93.8\textsubscript{0.6}} \\
         + \textsc{bt-tapt} & \textbf{94.4\textsubscript{0.1}} & \textbf{67.7\textsubscript{0.8}} & \textbf{92.6\textsubscript{0.1}} & 96.5\textsubscript{0.3} & \textbf{89.7\textsubscript{0.3}} & \textbf{93.8\textsubscript{0.5}} \\
    \bottomrule
    \end{tabular}{\parfillskip=0pt\par}
    \caption{Average classification performance with the standard deviation (subscripted) for each test dataset with three benchmark methods with two base LMs under a low-resource setting. Note that the performance metric for \textsc{Amazon} dataset is macro-$F_1$ measure while that of the other datasets are the simple accuracy.}
    \label{tab:2}
\end{table*}

%% file: tables/3.exp_high.tex
\begin{table}[t]
    \centering
    \begin{tabular}{lccc}
    \toprule 
         & \small $^\dagger$IMDB & \small $^\dagger$\textsc{Amazon} & \small $^\dagger$\textsc{AGNews}  \\ \hline
         \small BERT& \small 93.7\textsubscript{0.1} & \small 65.3\textsubscript{1.4} & \small 94.1\textsubscript{0.1} \\
         \small + \textsc{tapt} & \small 94.8\textsubscript{0.2} & \small 68.1\textsubscript{2.8} & \small \textbf{94.7\textsubscript{0.2}} \\
         \small + \textsc{bt-tapt} & \small \textbf{95.1\textsubscript{0.1}} & \small \textbf{69.7\textsubscript{1.9}} & \small 94.6\textsubscript{0.1} \\ \hline
         \small \textsc{RoBERTa}\xspace & \small 95.1\textsubscript{0.1} & \small 65.7\textsubscript{2.2} & \small 94.6\textsubscript{0.3} \\
         \small + \textsc{tapt} & \small 95.6\textsubscript{0.1} & \small 68.5\textsubscript{1.2} & \small 94.9\textsubscript{0.1} \\
         \small + \textsc{bt-tapt} & \small \textbf{95.7\textsubscript{0.0}} & \small \textbf{69.2\textsubscript{0.9}} & \small \textbf{95.0\textsubscript{0.1}} \\ 
    \bottomrule
    \end{tabular}{\parfillskip=0pt\par}
    \caption{Average classification performance with the standard deviation (subscripted) for each test dataset with three benchmark methods with two base LMs under a high-resource setting. The performance metrics are same with Table \ref{tab:2}.  $^\dagger$ denotes a high-resource setting where the entire dataset is used.}
    \label{tab:3}
\end{table}

%% file: tables/4.change_augmentation.tex
\begin{table}[t]
    \centering
    \begin{tabular}{cc}
    \toprule
         \small \textbf{Augmentation}& \small \textbf{Accuracy}\\
         \midrule
         \small TAPT& \small 93.0\textsubscript{0.2}\\ 
         \midrule
         \small None& \small 92.8\textsubscript{0.2} ($\downarrow$0.2\%p)\\
         \small + EDA& \small 92.7\textsubscript{0.2} ($\downarrow$0.3\%p)\\
         \small + Embedding& \small 92.9\textsubscript{0.2} ($\downarrow$0.1\%p)\\
         \small + TF-IDF& \small 92.9\textsubscript{0.4} ($\downarrow$0.1\%p)\\
         \small + \textbf{Back-Translation}& \small \textbf{93.3\textsubscript{0.2} ($\uparrow$0.3\%p)}\\ 
    \bottomrule
    \end{tabular}{\parfillskip=0pt\par}
    \caption{The average performance of the fine-tuned BERT-base model on IMDB low-resource test dataset with further re-pretraining using additional data generated from various augmentation technique after \textbf{TAPT} is applied.}
    \label{tab:4}
\end{table}

%% file: 5_robustness_noise.tex
\begin{figure*}[t!]
\centering{\includegraphics[width=0.965\textwidth]{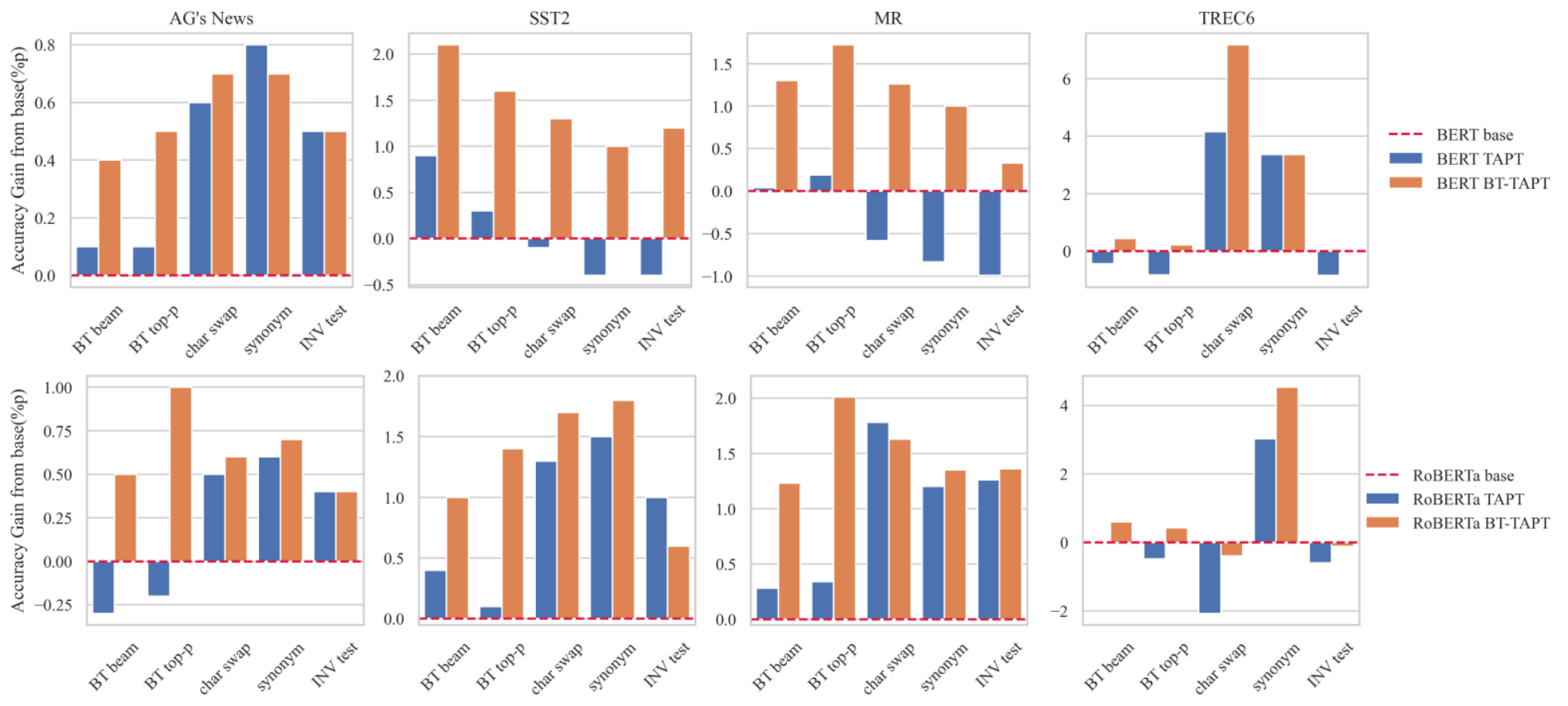}}
\caption
{Accuracy gains of BERT and \textsc{RoBERTa}\xspace on four noise-added test datasets when \textbf{TAPT} and \textbf{BT-TAPT} were applied.}
\label{fig:5}
\centering
\end{figure*}

\section{Robustness to Noise}

When applying the fine-tuned classifier to a real task, not only the clean data, which were sufficiently observed during the model training, but also the noisy data often come during the inference. We expect the proposed \textbf{TP-TAPT} to be more robust to various noises because the model trained with \textbf{BT-TAPT} encounters more diverse contexts than \textbf{TAPT}. To verify this assumption, we constructed another noisy-test dataset by applying corruption or perturbation and compared the classification performances of the three models. 

\subsection{Type of Noises}
Five different realistic noise types were generated for the test datasets \textsc{AGnews}, SST2, MR, and TREC6. Note that these noises are added only to the test dataset, not the training dataset.
\para{Synonym} Replace the random word in the sentence with its synonyms using WordNet thesaurus (replacement probability $=0.1$).
\para{BT beam} Generate a paraphrase using back-translation with beam search. This method does not shift the original sentence substantially because the beam search does not employ sampling. Nevertheless, it produces more changes than the synonym replacement.
\para{BT top-p} Generate multiple paraphrases using back-translation with nucleus sampling. It creates diverse but partially incorrect sentences.
\para{Char swap} Generate random character-level noises with transformation probabilities of 0.1. The noise consists of deleting, adding, or changing the order of characters in a sentence.
\para{InvTest} Apply the invariance test proposed in \citet{ribeiro2020beyond}. The invariance test applies label preserving-perturbations such as changing numbers or location names. The fine-tuned classifier should generate the same output whether the input changes by the invariance test.

Among the five noises scenarios, \textbf{BT beam} and \textbf{BT top-p} used English $\rightarrow$ German and German $\rightarrow$ English machine translation model adopted from \citet{ng2019facebook}, which is the same model used in section \ref{sec:exp}. CharSwap, Synonym, and InvTest used the TextAttack \cite{morris2020textattack} python package. As randomness exists in all methods except for the BT beam, where the maximum probability is used for decoding, we created five different noised datasets for each dataset.

\subsection{Result}
To evaluate the robustness of \textbf{TAPT} and \textbf{BT-TAPT}, we measured the \textit{accuracy gain}, the amount of classification accuracy change on the noised test set after \textbf{TAPT} or \textbf{BT-TAPT} was applied. Figure \ref{fig:5} shows the \textit{accuracy gain} regarding each dataset under different noise-added settings. Note that because the first two noise types, i.e., \textbf{BT beam} and \textbf{BT top-p} can generate similar sentences to \textbf{BT-TAPT}, \textbf{BT-TAPT} always improves the base re-pretrained model with a significant margin for most datasets, whereas \textbf{TAPT} sometimes fails to improve the base model performances. Beside these back-translation-based noises, \textbf{BT-TAPT} still reports favorable accuracy gains. However, \textbf{TAPT} not only achieves less significant accuracy gain than \textbf{BT-TAPT}, but also it degenerates the pretrained model (negative accuracy gain) even though it re-pretrained the base LMs. Consequently, we can conclude that the proposed \textbf{BT-TAPT} is more robust to unexpected data variations. Thus, the proposed method will be practically helpful in real-world tasks where languages are dynamically changing, evolving, and sometimes intentionally or unintentionally breaking.

%% file: 7_conclusion.tex
\section{Conclusion}

In this paper, we proposed \textbf{BT-TAPT}, a new adaptive pretraining method for generalizing LMs to task domains when task data is insufficient. In contrast to \textbf{TAPT} that only uses task-related unlabeled data, the proposed \textbf{BT-TAPT} generates augmentated data based on back-translation and use it for further re-pretrain the LMs. 

Experiments on six text classification datasets show that the \textbf{BT-TAPT} not only improved the classification accuracy but also reduced the deviation. Moreover, \textbf{BT-TAPT} was found to be more practical for small datasets and robust to various noises. Future work will consider applying \textbf{BT-TAPT} to the other NLP tasks such as question answering or summarization.

%% file: 00_main.bbl
\begin{thebibliography}{38}
\expandafter\ifx\csname natexlab\endcsname\relax\def\natexlab#1{#1}\fi

\bibitem[{Anaby-Tavor et~al.(2020)Anaby-Tavor, Carmeli, Goldbraich, Kantor,
  Kour, Shlomov, Tepper, and Zwerdling}]{anaby2020not}
Ateret Anaby-Tavor, Boaz Carmeli, Esther Goldbraich, Amir Kantor, George Kour,
  Segev Shlomov, Naama Tepper, and Naama Zwerdling. 2020.
\newblock Do not have enough data? deep learning to the rescue!
\newblock In \emph{Proceedings of the AAAI Conference on Artificial
  Intelligence}, volume~34, pages 7383--7390.

\bibitem[{Beltagy et~al.(2019)Beltagy, Lo, and Cohan}]{beltagy2019scibert}
Iz~Beltagy, Kyle Lo, and Arman Cohan. 2019.
\newblock Scibert: A pretrained language model for scientific text.
\newblock In \emph{Proceedings of the 2019 Conference on Empirical Methods in
  Natural Language Processing and the 9th International Joint Conference on
  Natural Language Processing (EMNLP-IJCNLP)}, pages 3606--3611.

\bibitem[{Brown et~al.(2020)Brown, Mann, Ryder, Subbiah, Kaplan, Dhariwal,
  Neelakantan, Shyam, Sastry, Askell et~al.}]{brown2020language}
Tom~B Brown, Benjamin Mann, Nick Ryder, Melanie Subbiah, Jared Kaplan, Prafulla
  Dhariwal, Arvind Neelakantan, Pranav Shyam, Girish Sastry, Amanda Askell,
  et~al. 2020.
\newblock Language models are few-shot learners.
\newblock \emph{arXiv preprint arXiv:2005.14165}.

\bibitem[{Devlin et~al.(2019)Devlin, Chang, Lee, and
  Toutanova}]{devlin2019bert}
Jacob Devlin, Ming-Wei Chang, Kenton Lee, and Kristina Toutanova. 2019.
\newblock Bert: Pre-training of deep bidirectional transformers for language
  understanding.
\newblock In \emph{Proceedings of the 2019 Conference of the North American
  Chapter of the Association for Computational Linguistics: Human Language
  Technologies, Volume 1 (Long and Short Papers)}, pages 4171--4186.

\bibitem[{Edunov et~al.(2018)Edunov, Ott, Auli, and
  Grangier}]{edunov2018understanding}
Sergey Edunov, Myle Ott, Michael Auli, and David Grangier. 2018.
\newblock Understanding back-translation at scale.
\newblock In \emph{Proceedings of the 2018 Conference on Empirical Methods in
  Natural Language Processing}, pages 489--500.

\bibitem[{Gokaslan and Cohen(2019)}]{Gokaslan2019OpenWeb}
Aaron Gokaslan and Vanya Cohen. 2019.
\newblock Openwebtext corpus.
\newblock \url{http://Skylion007.github.io/OpenWebTextCorpus}.

\bibitem[{Gururangan et~al.(2020)Gururangan, Marasovi{\'c}, Swayamdipta, Lo,
  Beltagy, Downey, and Smith}]{gururangan2020don}
Suchin Gururangan, Ana Marasovi{\'c}, Swabha Swayamdipta, Kyle Lo, Iz~Beltagy,
  Doug Downey, and Noah~A Smith. 2020.
\newblock Don't stop pretraining: Adapt language models to domains and tasks.
\newblock \emph{arXiv preprint arXiv:2004.10964}.

\bibitem[{He and McAuley(2016)}]{he2016ups}
Ruining He and Julian McAuley. 2016.
\newblock Ups and downs: Modeling the visual evolution of fashion trends with
  one-class collaborative filtering.
\newblock In \emph{proceedings of the 25th international conference on world
  wide web}, pages 507--517.

\bibitem[{Holtzman et~al.(2019)Holtzman, Buys, Du, Forbes, and
  Choi}]{holtzman2019curious}
Ari Holtzman, Jan Buys, Li~Du, Maxwell Forbes, and Yejin Choi. 2019.
\newblock The curious case of neural text degeneration.
\newblock In \emph{International Conference on Learning Representations}.

\bibitem[{Howard and Ruder(2018)}]{howard2018universal}
Jeremy Howard and Sebastian Ruder. 2018.
\newblock Universal language model fine-tuning for text classification.
\newblock In \emph{Proceedings of the 56th Annual Meeting of the Association
  for Computational Linguistics (Volume 1: Long Papers)}, pages 328--339.

\bibitem[{Kumar et~al.(2020)Kumar, Choudhary, and Cho}]{kumar2020data}
Varun Kumar, Ashutosh Choudhary, and Eunah Cho. 2020.
\newblock Data augmentation using pre-trained transformer models.
\newblock In \emph{Proceedings of the 2nd Workshop on Life-long Learning for
  Spoken Language Systems}, pages 18--26.

\bibitem[{Lan et~al.(2019)Lan, Chen, Goodman, Gimpel, Sharma, and
  Soricut}]{lan2019albert}
Zhenzhong Lan, Mingda Chen, Sebastian Goodman, Kevin Gimpel, Piyush Sharma, and
  Radu Soricut. 2019.
\newblock Albert: A lite bert for self-supervised learning of language
  representations.
\newblock In \emph{International Conference on Learning Representations}.

\bibitem[{Lee et~al.(2020)Lee, Yoon, Kim, Kim, Kim, So, and
  Kang}]{lee2020biobert}
Jinhyuk Lee, Wonjin Yoon, Sungdong Kim, Donghyeon Kim, Sunkyu Kim, Chan So, and
  Jaewoo Kang. 2020.
\newblock Biobert: a pre-trained biomedical language representation model for
  biomedical text mining.
\newblock \emph{Bioinformatics}, 36(4):1234--1240.

\bibitem[{Lewis et~al.(2020)Lewis, Liu, Goyal, Ghazvininejad, Mohamed, Levy,
  Stoyanov, and Zettlemoyer}]{lewis-etal-2020-bart}
Mike Lewis, Yinhan Liu, Naman Goyal, Marjan Ghazvininejad, Abdelrahman Mohamed,
  Omer Levy, Veselin Stoyanov, and Luke Zettlemoyer. 2020.
\newblock \href {https://doi.org/10.18653/v1/2020.acl-main.703} {{BART}:
  Denoising sequence-to-sequence pre-training for natural language generation,
  translation, and comprehension}.
\newblock In \emph{Proceedings of the 58th Annual Meeting of the Association
  for Computational Linguistics}, pages 7871--7880, Online. Association for
  Computational Linguistics.

\bibitem[{Li and Roth(2002)}]{li2002learning}
Xin Li and Dan Roth. 2002.
\newblock Learning question classifiers.
\newblock In \emph{COLING 2002: The 19th International Conference on
  Computational Linguistics}.

\bibitem[{Liu et~al.(2016)Liu, Qiu, and Huang}]{liu2016recurrent}
Pengfei Liu, Xipeng Qiu, and Xuanjing Huang. 2016.
\newblock Recurrent neural network for text classification with multi-task
  learning.
\newblock In \emph{Proceedings of the Twenty-Fifth International Joint
  Conference on Artificial Intelligence}, pages 2873--2879.

\bibitem[{Liu et~al.(2019)Liu, Ott, Goyal, Du, Joshi, Chen, Levy, Lewis,
  Zettlemoyer, and Stoyanov}]{liu2019roberta}
Yinhan Liu, Myle Ott, Naman Goyal, Jingfei Du, Mandar Joshi, Danqi Chen, Omer
  Levy, Mike Lewis, Luke Zettlemoyer, and Veselin Stoyanov. 2019.
\newblock Roberta: A robustly optimized bert pretraining approach.
\newblock \emph{arXiv preprint arXiv:1907.11692}.

\bibitem[{Maas et~al.(2011)Maas, Daly, Pham, Huang, Ng, and
  Potts}]{maas2011learning}
Andrew Maas, Raymond~E Daly, Peter~T Pham, Dan Huang, Andrew~Y Ng, and
  Christopher Potts. 2011.
\newblock Learning word vectors for sentiment analysis.
\newblock In \emph{Proceedings of the 49th annual meeting of the association
  for computational linguistics: Human language technologies}, pages 142--150.

\bibitem[{Morris et~al.(2020)Morris, Lifland, Yoo, and
  Qi}]{morris2020textattack}
John~X Morris, Eli Lifland, Jin~Yong Yoo, and Yanjun Qi. 2020.
\newblock Textattack: A framework for adversarial attacks in natural language
  processing.
\newblock \emph{arXiv preprint arXiv:2005.05909}.

\bibitem[{Mrk{\v{s}}i{\'c} et~al.(2016)Mrk{\v{s}}i{\'c}, S{\'e}aghdha, Thomson,
  Gasic, Barahona, Su, Vandyke, Wen, and Young}]{mrkvsic2016counter}
Nikola Mrk{\v{s}}i{\'c}, Diarmuid~{\'O} S{\'e}aghdha, Blaise Thomson, Milica
  Gasic, Lina M~Rojas Barahona, Pei-Hao Su, David Vandyke, Tsung-Hsien Wen, and
  Steve Young. 2016.
\newblock Counter-fitting word vectors to linguistic constraints.
\newblock In \emph{Proceedings of the 2016 Conference of the North American
  Chapter of the Association for Computational Linguistics: Human Language
  Technologies}, pages 142--148.

\bibitem[{Ng et~al.(2019)Ng, Yee, Baevski, Ott, Auli, and
  Edunov}]{ng2019facebook}
Nathan Ng, Kyra Yee, Alexei Baevski, Myle Ott, Michael Auli, and Sergey Edunov.
  2019.
\newblock Facebook fair’s wmt19 news translation task submission.
\newblock In \emph{Proceedings of the Fourth Conference on Machine Translation
  (Volume 2: Shared Task Papers, Day 1)}, pages 314--319.

\bibitem[{Pang and Lee(2005)}]{pang2005seeing}
Bo~Pang and Lillian Lee. 2005.
\newblock Seeing stars: Exploiting class relationships for sentiment
  categorization with respect to rating scales.
\newblock In \emph{Proceedings of the 43rd Annual Meeting of the Association
  for Computational Linguistics (ACL’05)}, pages 115--124.

\bibitem[{Peters et~al.(2018)Peters, Neumann, Iyyer, Gardner, Clark, Lee, and
  Zettlemoyer}]{peters2018deep}
Matthew~E Peters, Mark Neumann, Mohit Iyyer, Matt Gardner, Christopher Clark,
  Kenton Lee, and Luke Zettlemoyer. 2018.
\newblock Deep contextualized word representations.
\newblock In \emph{Proceedings of NAACL-HLT}, pages 2227--2237.

\bibitem[{Raffel et~al.(2020)Raffel, Shazeer, Roberts, Lee, Narang, Matena,
  Zhou, Li, and Liu}]{raffel2020exploring}
Colin Raffel, Noam Shazeer, Adam Roberts, Katherine Lee, Sharan Narang, Michael
  Matena, Yanqi Zhou, Wei Li, and Peter~J Liu. 2020.
\newblock Exploring the limits of transfer learning with a unified text-to-text
  transformer.
\newblock \emph{Journal of Machine Learning Research}, 21:1--67.

\bibitem[{Ribeiro et~al.(2020)Ribeiro, Wu, Guestrin, and
  Singh}]{ribeiro2020beyond}
Marco~Tulio Ribeiro, Tongshuang Wu, Carlos Guestrin, and Sameer Singh. 2020.
\newblock Beyond accuracy: Behavioral testing of nlp models with checklist.
\newblock \emph{arXiv preprint arXiv:2005.04118}.

\bibitem[{Sennrich et~al.(2016{\natexlab{a}})Sennrich, Haddow, and
  Birch}]{sennrich2016improving}
Rico Sennrich, Barry Haddow, and Alexandra Birch. 2016{\natexlab{a}}.
\newblock Improving neural machine translation models with monolingual data.
\newblock In \emph{Proceedings of the 54th Annual Meeting of the Association
  for Computational Linguistics (Volume 1: Long Papers)}, pages 86--96.

\bibitem[{Sennrich et~al.(2016{\natexlab{b}})Sennrich, Haddow, and
  Birch}]{sennrich2016neural}
Rico Sennrich, Barry Haddow, and Alexandra Birch. 2016{\natexlab{b}}.
\newblock Neural machine translation of rare words with subword units.
\newblock In \emph{Proceedings of the 54th Annual Meeting of the Association
  for Computational Linguistics (Volume 1: Long Papers)}, pages 1715--1725.

\bibitem[{Socher et~al.(2013)Socher, Bauer, Manning, and
  Ng}]{socher2013parsing}
Richard Socher, John Bauer, Christopher~D Manning, and Andrew~Y Ng. 2013.
\newblock Parsing with compositional vector grammars.
\newblock In \emph{Proceedings of the 51st Annual Meeting of the Association
  for Computational Linguistics (Volume 1: Long Papers)}, pages 455--465.

\bibitem[{Song et~al.(2019)Song, Tan, Qin, Lu, and Liu}]{song2019mass}
Kaitao Song, Xu~Tan, Tao Qin, Jianfeng Lu, and Tie-Yan Liu. 2019.
\newblock Mass: Masked sequence to sequence pre-training for language
  generation.
\newblock In \emph{International Conference on Machine Learning}, pages
  5926--5936. PMLR.

\bibitem[{Sun et~al.(2019)Sun, Qiu, Xu, and Huang}]{sun2019fine}
Chi Sun, Xipeng Qiu, Yige Xu, and Xuanjing Huang. 2019.
\newblock How to fine-tune bert for text classification?
\newblock In \emph{China National Conference on Chinese Computational
  Linguistics}, pages 194--206. Springer.

\bibitem[{Trinh and Le(2018)}]{trinh2018simple}
Trieu~H Trinh and Quoc~V Le. 2018.
\newblock A simple method for commonsense reasoning.
\newblock \emph{arXiv preprint arXiv:1806.02847}.

\bibitem[{Wang and Yang(2015)}]{wang2015s}
William~Yang Wang and Diyi Yang. 2015.
\newblock That’s so annoying!!!: A lexical and frame-semantic embedding based
  data augmentation approach to automatic categorization of annoying behaviors
  using\# petpeeve tweets.
\newblock In \emph{Proceedings of the 2015 conference on empirical methods in
  natural language processing}, pages 2557--2563.

\bibitem[{Wei and Zou(2019)}]{wei2019eda}
Jason Wei and Kai Zou. 2019.
\newblock Eda: Easy data augmentation techniques for boosting performance on
  text classification tasks.
\newblock In \emph{Proceedings of the 2019 Conference on Empirical Methods in
  Natural Language Processing and the 9th International Joint Conference on
  Natural Language Processing (EMNLP-IJCNLP)}, pages 6383--6389.

\bibitem[{Wolf et~al.(2020)Wolf, Debut, Sanh, Chaumond, Delangue, Moi, Cistac,
  Rault, Louf, Funtowicz, Davison, Shleifer, von Platen, Ma, Jernite, Plu, Xu,
  Scao, Gugger, Drame, Lhoest, and Rush}]{wolf-etal-2020-transformers}
Thomas Wolf, Lysandre Debut, Victor Sanh, Julien Chaumond, Clement Delangue,
  Anthony Moi, Pierric Cistac, Tim Rault, Rémi Louf, Morgan Funtowicz, Joe
  Davison, Sam Shleifer, Patrick von Platen, Clara Ma, Yacine Jernite, Julien
  Plu, Canwen Xu, Teven~Le Scao, Sylvain Gugger, Mariama Drame, Quentin Lhoest,
  and Alexander~M. Rush. 2020.
\newblock \href {https://www.aclweb.org/anthology/2020.emnlp-demos.6}
  {Transformers: State-of-the-art natural language processing}.
\newblock In \emph{Proceedings of the 2020 Conference on Empirical Methods in
  Natural Language Processing: System Demonstrations}, pages 38--45, Online.
  Association for Computational Linguistics.

\bibitem[{Xie et~al.(2019)Xie, Dai, Hovy, Luong, and Le}]{xie2019unsupervised}
Qizhe Xie, Zihang Dai, Eduard Hovy, Minh-Thang Luong, and Quoc~V Le. 2019.
\newblock Unsupervised data augmentation for consistency training.
\newblock \emph{arXiv preprint arXiv:1904.12848}.

\bibitem[{Yu et~al.(2018)Yu, Dohan, Luong, Zhao, Chen, Norouzi, and
  Le}]{yu2018qanet}
Adams~Wei Yu, David Dohan, Minh-Thang Luong, Rui Zhao, Kai Chen, Mohammad
  Norouzi, and Quoc~V Le. 2018.
\newblock Qanet: Combining local convolution with global self-attention for
  reading comprehension.
\newblock In \emph{International Conference on Learning Representations}.

\bibitem[{Zhang et~al.(2015)Zhang, Zhao, and LeCun}]{zhang2015character}
Xiang Zhang, Junbo Zhao, and Yann LeCun. 2015.
\newblock Character-level convolutional networks for text classification.
\newblock \emph{Advances in neural information processing systems},
  28:649--657.

\bibitem[{Zhu et~al.(2015)Zhu, Kiros, Zemel, Salakhutdinov, Urtasun, Torralba,
  and Fidler}]{zhu2015aligning}
Yukun Zhu, Ryan Kiros, Rich Zemel, Ruslan Salakhutdinov, Raquel Urtasun,
  Antonio Torralba, and Sanja Fidler. 2015.
\newblock Aligning books and movies: Towards story-like visual explanations by
  watching movies and reading books.
\newblock In \emph{Proceedings of the IEEE international conference on computer
  vision}, pages 19--27.

\end{thebibliography}
